\documentclass{article}
\usepackage{spconf,amsmath,graphicx}
\usepackage{amsfonts}
\usepackage{amssymb}
\usepackage{amsthm}
\usepackage{mathrsfs}
\usepackage{multirow}
\usepackage{multicol} 
\usepackage{arydshln}
\usepackage[table,xcdraw]{xcolor}
\usepackage{bigstrut}
\usepackage{bbding}
\usepackage{booktabs}
\usepackage{stfloats}

\title{open-world object detection via discriminative class prototype learning}
\name{Jinan Yu$^{1,2}$ \qquad Liyan Ma$^{1,2}$\sthanks{Corresponding authors, \{liyanma, zhenglin\_li\}@shu.edu.cn. This work was supported in part by the Shanghai Municipal Natural Science Foundation under Grant 21ZR1423300, in part by the National Key R\&D Program of China (No. 2021YFA1003004), in part by Shanghai Sailing Program under Grant 22YF1413800.} \qquad Zhenglin Li$^{2*}$ \qquad Yan Peng$^2$ \qquad Shaorong Xie$^{1,2}$}
\address{$^{1}$ School of Computer Engineering and Science, Shanghai University, Shanghai, China\\
      $^{2}$ School of Artificial Intellegence, Shanghai University,Shanghai, China}
\begin{document}
%
\maketitle
\begin{abstract}
Open-world object detection (OWOD) is a challenging problem that combines object detection with incremental learning and open-set learning. Compared to standard object detection, the OWOD setting is task to: 1) detect objects seen during training while identifying unseen classes, and 2) incrementally learn the knowledge of the identified unknown objects when the corresponding annotations is available. We propose a novel and efficient OWOD solution from a prototype perspective, which we call OCPL: Open-world object detection via discriminative Class Prototype Learning, which consists of a Proposal Embedding Aggregator (PEA), an Embedding Space Compressor (ESC) and a Cosine Similarity-based Classifier (CSC). All our proposed modules aim to learn the discriminative embeddings of known classes in the feature space to minimize the overlapping distributions of known and unknown classes, which is beneficial to differentiate known and unknown classes. Extensive experiments performed on PASCAL VOC and MS-COCO benchmark demonstrate the effectiveness of our proposed method.
\end{abstract}
\section{Introduction}
\label{sec:intro}
The rapid development of deep learning in recent years has significantly improved the performance of object detection, which is to identify and localize regions of interest in an image. But almost all existing object detectors consider the close-world assumption that the test sets and training sets contain the same data categories \cite{redmon2016you,lin2017focal,ren2015faster,tian2019fcos}. In practical applications, the test set may emerge classes not seen in the training set. Therefore, 
open-world object detection was proposed to solve the challenge that the model not only needs to correctly detect known classes, but also recognize unknown classes \cite{joseph2021towards}. Furthermore the detector progressively learn the new knowledge once identified unknown categories are annotated. 

The identification of unknown classes in traditional object detection pipeline faces significant challenges. Dhamija et al. were the first to formalize the open-set object detection problem \cite{dhamija2020overlooked}. They revealed that even State-of-the-art object detectors also result in false positive detection in the open-set environment. Complementing the benchmark for open-set object detection, Joseph et al. proposed the first open-world object detection model, named ORE \cite{joseph2021towards}, based on the Faster-RCNN \cite{ren2015faster}. Since the annotations of unknown object are unavailable, ORE introduces the auto-labelling unknowns module to obtain a weakly supervised set of unknown objects. 

Although ORE was the first to introduce and explore the challenging OWOD paradigm, it suffers from several defects. 1) The effect of the auto-labelling mechanism is negligible. It is difficult for the model to obtain supervision of unknown samples from this module and may cause confusion between background and unknown classes.  2) Training an energy-based classifier in ORE requires a fully annotated dataset, which contains label information of known and unknown classes. Obviously, ORE violates the principle of open-world object detection, where only annotations of known classes are available during training. 

Motivated by the above observations, we intend to build an open-world object detection pipeline that does not require pseudo-supervised for unknown classes and illogical extra datasets for unknown classes. In fact, Region Proposal Network (RPN) can extract proposals with high objectness scores, which contain both known and unknown classes. In other words, If the overlap of the distributions of unknown and known classes can be largely avoided, we can differentiate them well. Inspired by work on open set recognition \cite{yang2018robust,miller2021class,yang2020convolutional,chen2021adversarial,xia2021spatial,chen2020learning}, we introduce a prototype branch to cluster known classes. Proposal Embedding Aggregator (PEA) and Embedding Space Compressor (ESC) are proposed to make compact clusters. Furthermore, we also introduce a 
Cosine Similarity-Based classifier (CSC) to help form tighter clusters \cite{wang2018cosface}. We summarize our contributions as follows:
\begin{figure*}[htb]
\centering
\centerline{\includegraphics[width=0.9\linewidth]{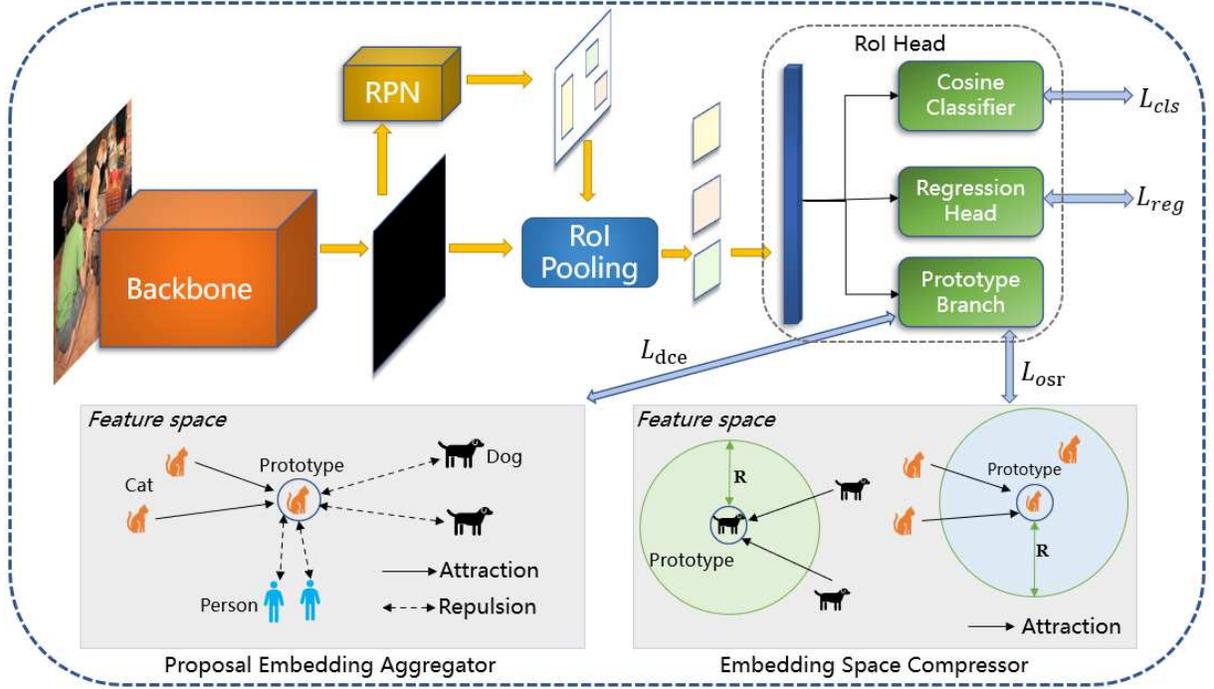}}
\caption{An overview of proposed OCPL. (\texttt{top}) We introduce a prototype branch into Faster-RCNN and adopt a Cosine Classifier. (\texttt{bottom}) Proposal Embedding Aggregator and Embedding Space Compressor are proposed in prototype branch.}
\label{fig:network}
\end{figure*}

1. We are the first to explore the possibility of prototype ideology in open-world object detection.

2. We introduce a prototype branch with Proposal Embedding Aggregator and Embedding Space Compressor which are adopted to separate and squeeze the feature distribution of each known class respectively. In order to form more compact class clusters, we introduce a Cosine Similarity-based Classifier.

3. Extensive experiments on the OWOD benchmark confirm that our proposed OCPL is satisfactory. Specifically, the proposed method outperforms the recently proposed ORE on most metrics.
\section{METHOD}
\label{sec:method}

\subsection{Overall Architecture}

To integrate prototype learning with object detection frameworks, we choose the classic two-stage Faster-RCNN framework. Faster-RCNN can sample proposals of different sizes into the same shape through RoI pooling, which means it is beneficial for prototype learning. Fig.1 shows the overall structure of the proposed OCPL. In the first stage, the feature maps output of the backone are fed into a class-agnostic Region Proposal Network (RPN) to propose potential regions that may have an object. The next stage model classifies, regresses and learns representations for each proposed region through three parallel fully connected layers. In the prototype branch, we learn the distribution of known classes via Proposal Embedding Aggregator and Embedding Space Compressor. Then the distance between the features of the sample and the prototype centers can be used to measure the probability that the sample belongs to a known or unknown class. To further form tighter clusters, we replace the original classifier with a Cosine Similarity-based Classifier. The regression head and the corresponding loss $L_{reg}$ remain unchanged.

\subsection{Proposal Embedding Aggregator}

Some researchers \cite{yang2020convolutional} have shown that setting multiple prototypes for each cluster may be harmful to form tighter distribution of samples in the feature space, Therefore, only one prototype center is chosen for each class and initialized by one-hot encoding. We denote the prototype centers as $\mathcal{C} = \{ \mathcal{C}^k  \in \mathbb{R}^d ,k = 1,2,...,K\}$, where $K$ is number of categories, and we denote $\mathcal{F}$ as the function of RoI Head. For any instance $x_i  \in \mathbb{R}^{C \times H \times W} $ from RoI pooling, we adopt the setting of ARPL \cite{chen2021adversarial} to measure the distance between $x_i$ and the prototype center of category $j$:
\[
\begin{gathered}
  \mathcal{D}\left( {\mathcal{F}\left( {x_i } \right)_p ,\mathcal{C}^j } \right) = \mathcal{D}_e \left( {\mathcal{F}\left( {x_i } \right)_p ,\mathcal{C}^j } \right) - \mathcal{D}_d \left( {\mathcal{F}\left( {x_i } \right)_p ,\mathcal{C}^j } \right), \hfill \\\tag{1}
  \mathcal{D}_e \left( {\mathcal{F}\left( {x_i } \right)_p ,\mathcal{C}^j } \right) = \frac{1}
  {d}\left\| {\mathcal{F}\left( {x_i } \right)_p  - \mathcal{C}^j } \right\|_2^2 , \hfill \\
  \mathcal{D}_d \left( {\mathcal{F}\left( {x_i } \right)_p ,\mathcal{C}^j  } \right) = \mathcal{F}\left( {x_i } \right)_p  \bullet \mathcal{C}^j  , \hfill \\ 
\end{gathered} 
\]
where $\mathcal{F}\left( {x_i } \right)_p  \in \mathbb{R}^d$ is the embedding features of the prototype branch, $\mathcal{D}_e$ and $\mathcal{D}_d$ represent Euclidean distance and dot product, respectively. In order to optimize the prototype more conveniently, we imitate the cross entropy mechanism and use distance-based cross entropy loss. Therefore, the probability that instance $x_i$ belongs to class $j$ can be as
\[
p_{ij} \left( {y = j\left| {x_i } \right.,\mathcal{F},\mathcal{C}} \right) = \frac{{e^{ - \mathcal{D}\left( {\mathcal{F}\left( {x_i } \right)_p,\mathcal{C}^j } \right)} }}
{{\sum\nolimits_{k = 1}^K {e^{ - \mathcal{D}\left( {\mathcal{F}\left( {x_i } \right)_p,\mathcal{C}^k } \right)} } }},\tag{2}
\]
where the corresponding objective function is
\[
L_{dce} \left( {x_i ,\theta,\mathcal{C}} \right) =  - log\ p_{ij} (y = j\left| {x_i ,\mathcal{F},\mathcal{C}} \right.),\tag{3}
\]
where $\theta$ is the model parameter. The optimization of $L_{dce}$ tends to converge only when $\mathcal{F}\left( {x_i } \right)_p$ and $\mathcal{C}^j$ are almost in one direction and $\mathcal{F}\left( {x_i } \right)_p$ and $\mathcal{C}^j$ are very close, as shown in Fig.1. By optimizing Eq.(3), the embedding features of the identical category are attracted each other by the prototype centers, and vice versa.

\subsection{Embedding Space Compressor}
In addition, We hope to further squeeze the distribution range of known classes in the feature space to reduce the open space risk loss, which is the degree of overlap between the distribution of unknown classes and known classes \cite{xia2021adversarial}. Specifically, the open space risk loss can be constrained by the radius of the prototype centers. The bottom right of Fig.1 describes the principle of $L_{osr}$, and it can be expressed as
\[
L_{osr} \left( {x_i ,\theta ,\mathcal{C},R} \right) = \max \left\{ {0,\mathcal{D}_e( {\mathcal{F}\left( {x_i } \right)_p,\mathcal{C}^j}) - R} \right\},\tag{4}
\]
where $x_i$ represents training samples with label $j$ from RoI pooling, and $R$ is a learnable parameter with an initial value of 0. Due to the nature of Euclidean norm, the value of $L_{osr}$ is always greater than zero in the initial stage of model training. With the continuous optimization of the network, $R$ will gradually increase and converge to a value while $\mathcal{D}_e({\mathcal{F}\left( {x_i } \right)_p ,\mathcal{C}^j)}$ will decrease accordingly. Therefore, $L_{osr}$ can assist $L_{dce}$ to strengthen the discriminative ability of the model. Finally, the embedding features of each positive proposal samples will fall into a corresponding hypersphere with $\mathcal{C}^k$ as the center and $R$ as the radius.

The overall training objective function of the prototype branch can be expressed as:
\[
L_{proto}  = L_{dce}  + \lambda L_{osr}, \tag{5}
\]
where $\lambda$ is the balance factor. Our model is insensitive to $\lambda$ and we set $\lambda$ to 0.1 in all experiments.
\subsection{Cosine Similarity-based Classifier}
The features generated by the model trained by softmax loss are not discriminative enough, and the intra-class compactness is not considered that may cause the intra-class distance to be greater than the inter-class distance \cite{yang2020convolutional}. Therefore we use a cosine similarity-based classifier, where the logits before the softmax function of the classification loss is calculated by the scaled cosine similarity between instance features $f_i$ and class weights $W$,
\[
\\logits_i  = \alpha \frac{{W^T f_i }}
{{\left\| {W^T } \right\|\left\| {f_i } \right\|}},\tag{6}
\]
where $f_i  \in \mathbb{R}^C $ is obtained by global average pooling of $x_i$, and $\alpha$ is a scaled factor to strength gradient. Explicitly modeling similarity by cosine classifier \cite{wang2018cosface} helps to form tighter clustering of instances.

\textbf{During inference}, a threshold $\gamma $ is used to filter some detections with extremely low classification scores, usually $\gamma$ is set to 0.05. Then the output of the prototype layer is used to calculate the probability in Eq.(2) for the remaining samples. Finally, the class of instance $x_i$ can be determined by the following formula:
\[
class_i  = \left\{ {\begin{array}{*{20}c}
   j & , & {if{\kern 1pt} {\kern 1pt} {\kern 1pt} {\kern 1pt} {\kern 1pt} p_{ij}  \geqslant \xi ,}  \\
   {unknown} & , & {otherwise,} \tag{7} \\

 \end{array} } \right.
\]
where $p_{ij}$ is the probability that instance $x_i$ belongs to class $j$, and $\xi$ the threshold to decide whether $x_i$ is an unknown instance.

\begin{table*}[htbp]
  \centering
  \resizebox{\textwidth}{!}{
    \begin{tabular}{l|c|c|c|c|c|c|ccc|c|c|c|ccc|c|ccc}
    \toprule
    Task IDs & \multicolumn{4}{c|}{Task 1}   & \multicolumn{6}{c|}{Task 2}                   & \multicolumn{6}{c|}{Task 3}                   & \multicolumn{3}{c}{Task 4} \bigstrut\\
    \midrule
          & WI    & A-OSE & mAP(↑) & UR    & WI    & A-OSE & \multicolumn{3}{c|}{mAP(↑)} & UR    & WI    & A-OSE & \multicolumn{3}{c|}{mAP(↑)} & UR    & \multicolumn{3}{c}{mAP(↑)} \bigstrut\\
\cline{4-4}\cline{8-10}\cline{14-16}\cline{18-20}          & (↓)   & (↓)   & current & (↑)   & (↓)   & (↓)   & previous & current & both  & (↑)   & (↓)   & (↓)   & previous & current & both  & (↑)   & previous & current & both \bigstrut\\
    \hline
    Faster-RCNN & 0.0699 & 13396 & 56.16 & —     & 0.0371 & 12291 & 4.076 & 25.74 & 14.91 & —     & 0.0213 & 9174  & 6.96  & 13.48 & 9.138 & —     & 2.04 & 13.68 & 4.95 \bigstrut\\
    \hline
    \multicolumn{1}{p{6.25em}|}{Faster-RCNN\newline{}+Finetuning} & \multicolumn{4}{c|}{Not applicable as incremental} & 0.0375 & 12497 & 51.09 & 23.84 & 37.47 & —     & 0.0279 & 9622  & 35.69 & 11.53 & 27.64 & —     & 29.53 & 12.78 & 25.34 \bigstrut\\
    \hline
    ORE*  & 0.0531 & 12226 & 56.09 & 5.48  & 0.0319 & 10229 & {\bfseries51.80}  & 26.32 & 39.06 & 3.14  & 0.0192 & 8579  & 38.16 & 13.24 & 29.85 & 3.38  & 29.94 & 13.18 & 25.75 \bigstrut\\
    \hline
    \textbf{OCPL(ours)} & {\bfseries0.0423} & {\bfseries5670}  & {\bfseries56.64} & {\bfseries8.26}  & {\bfseries0.0220} & {\bfseries5690}  & 50.65 & {\bfseries27.54} & {\bfseries39.10}  & {\bfseries7.65}  & {\bfseries0.0162} & {\bfseries5166}  & {\bfseries38.63} & {\bfseries14.74} & {\bfseries30.67} & {\bfseries11.88} & {\bfseries30.75} & {\bfseries14.42} & {\bfseries26.67} \bigstrut\\
    \bottomrule
    \end{tabular}}%
  \caption{We show the performance of our proposed method on open-world object detection. (↑) means higher is better, and (↓) means lower is better. The "previous", "current" and "both" denote mAP of previously known classes, currently known classes and all known classes respectively. ORE* stands for ORE model without EBUI module (energy-based unknown identifier). }
  \label{tab:1}%
\end{table*}%

\begin{table}[htbp]
  \centering
  \resizebox{\linewidth}{!}{
    \begin{tabular}{c|c|ccc|c|c|c|c}
    \toprule
    Row ID & Prototype & PEA & ESC & CSC & WI & A-OSE & mAP & UR \bigstrut\\
    \midrule
    1     & learnable  &   \Checkmark    &   \XSolidBrush   &    \XSolidBrush    & 0.0442 & 5840 & 55.55 & 7.52 \bigstrut[t]\\
    2     & learnable  &   \Checkmark    &   \Checkmark     &    \XSolidBrush    & 0.0457 & 5866 & 54.78 & 7.74 \\
    3     & learnable  &   \Checkmark    &   \XSolidBrush   &    \Checkmark      & 0.0442 & 6140 & 55.84 & 7.71 \\
    4     & fixed+finetuning &   \Checkmark    &   \XSolidBrush   &    \XSolidBrush    & 0.0478 & 6155 & 56.12 & 6.65\\
    5     & fixed+finetuning &   \Checkmark    &   \Checkmark     &    \XSolidBrush    & 0.0431 & \textbf{5588} & 56.29 & 7.84 \\
    6     & fixed+finetuning &   \Checkmark    &   \Checkmark     &    \Checkmark      & \textbf{0.0423} & 5670 & \textbf{56.64} & \textbf{8.26} \bigstrut[b]\\
    \bottomrule
    \end{tabular}}%
  \label{tab:2}%
  \caption{Ablation study between different modules.}
\end{table}%

\section{EXPERIMENT}
\label{sec:experi}

\subsection{Experiment Setup}
Following data splitting in ORE, a total of 80 categories in the PASCAL VOC \cite{everingham2010pascal} and MSCOCO \cite{lin2014microsoft} training sets are equally divided into 4 groups to be trained in different tasks. We use Pascal VOC test split and MSCOCO val split for evaluation. For known classes, mAP still applies. For unknown classes, we adopt the Unknown Recall(UR) \cite{bansal2018zero} rate to represent the detection capability of the model for unknown objects. Wildness Impact (WI) \cite{dhamija2020overlooked} is used to measure the degree of confusion between unknown and known classes, and we use absolute Open-Set Error(A-OSE) \cite{miller2018dropout} to report the number of unknown objects detected as any known classes.

In each incremental learning task $T_i$, data in tasks after $T_i$ will be treated as unknown classes, only the training data for current task will be present in $T_i$. Therefore training without previous classes will cause catastrophic forgetting, which can be mitigated by fine-tuning network parameters with representative samples from previously known and currently known classes. Our model is based on the Faster-RCNN algorithm with a ResNet-50 \cite{he2016deep} backbone. The code is implemented in PyTorch  using Detectron2 \cite{wu2019detectron2}.

\begin{figure}[t]
\begin{minipage}[b]{0.49\linewidth}
  \centering
  \centerline{\includegraphics[width=\linewidth]{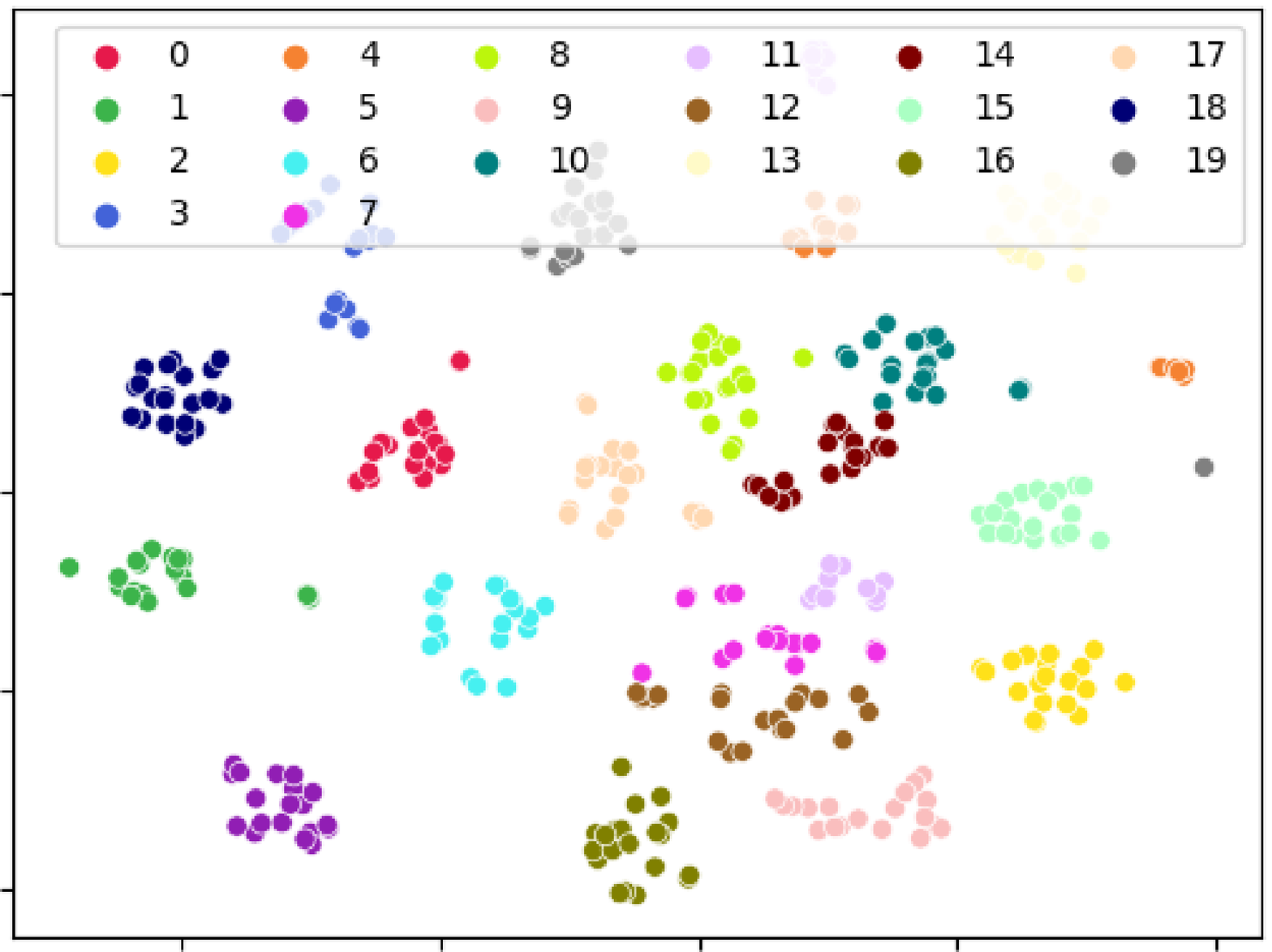}}
  \centerline{(a) ORE*}\medskip
\end{minipage}
\hfill
\begin{minipage}[b]{0.49\linewidth}
  \centering
  \centerline{\includegraphics[width=\linewidth]{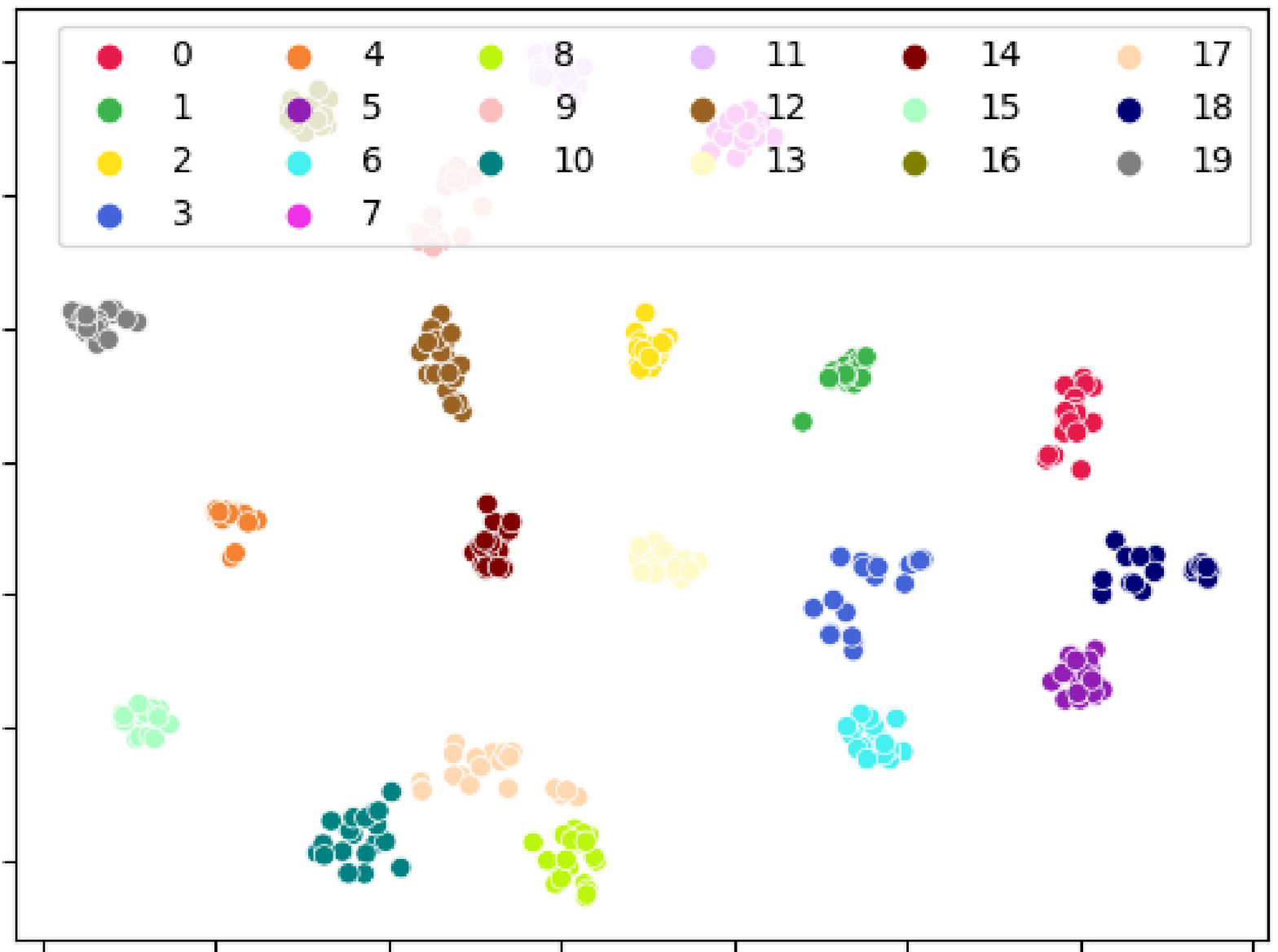}}
  \centerline{(b) Ours}\medskip
\end{minipage}
\caption{Comparison of feature visualization results}
\label{fig:t-SNE}
\end{figure}

\subsection{Experiment result}
Table 1 shows the comparison of our proposed OCPL model and other detectors under the OWOD evaluation protocol. It is obvious that the standard Faster-RCNN has no recall ability for unknown classes. To make a fair comparison with ORE, we reproduce the ORE* model, which is an ORE model without energy-based unknown identifier that relies on unknown dataset annotations. Our proposed model has more promising results than ORE* on almost all metrics.

The effect of each component we proposed was demonstrated in ablation experiments. The results are shown in Table 2. Two methods of prototype initialization are available. The "learnable" indicates that the prototype is randomly initialized and is learnable, whereas "fixed" means that each prototype center is initialized by one-hot encoding, which is more beneficial for modeling the class centers of complex datasets \cite{miller2021class}. To make the fixed prototype more flexible, we periodically fine-tune the prototype centers via the learned class features to accommodate data with large visual variation in each class. The results show that our fixed and fine-tuned prototype centers are more likely to generate compact and stable clusters than learnable prototype centers, and it is clear that R in Embedding Space Compressor(ESC) have difficulty adapting to changing prototype centers.
\subsection{Visualization and Analysis}
We visualized the quality of clusters of known classes trained in Task 1 by t-SNE \cite{van2008visualizing}, as shown in Fig.2. Our method forms more compact class clusters, which is beneficial for identifying unknown classes. Fig.3 presents the comparison of detection results. It's worth noting that '\texttt{cake}', '\texttt{pizza}' and '\texttt{elephant}' have not been trained in $T_1$. Our method successfully identifies unseen categories as '\texttt{unknown}', while ORE* is more inclined to false detections.
\begin{figure}[t]
\begin{minipage}[b]{0.49\linewidth}
  \centering
  \centerline{\includegraphics[width=\linewidth]{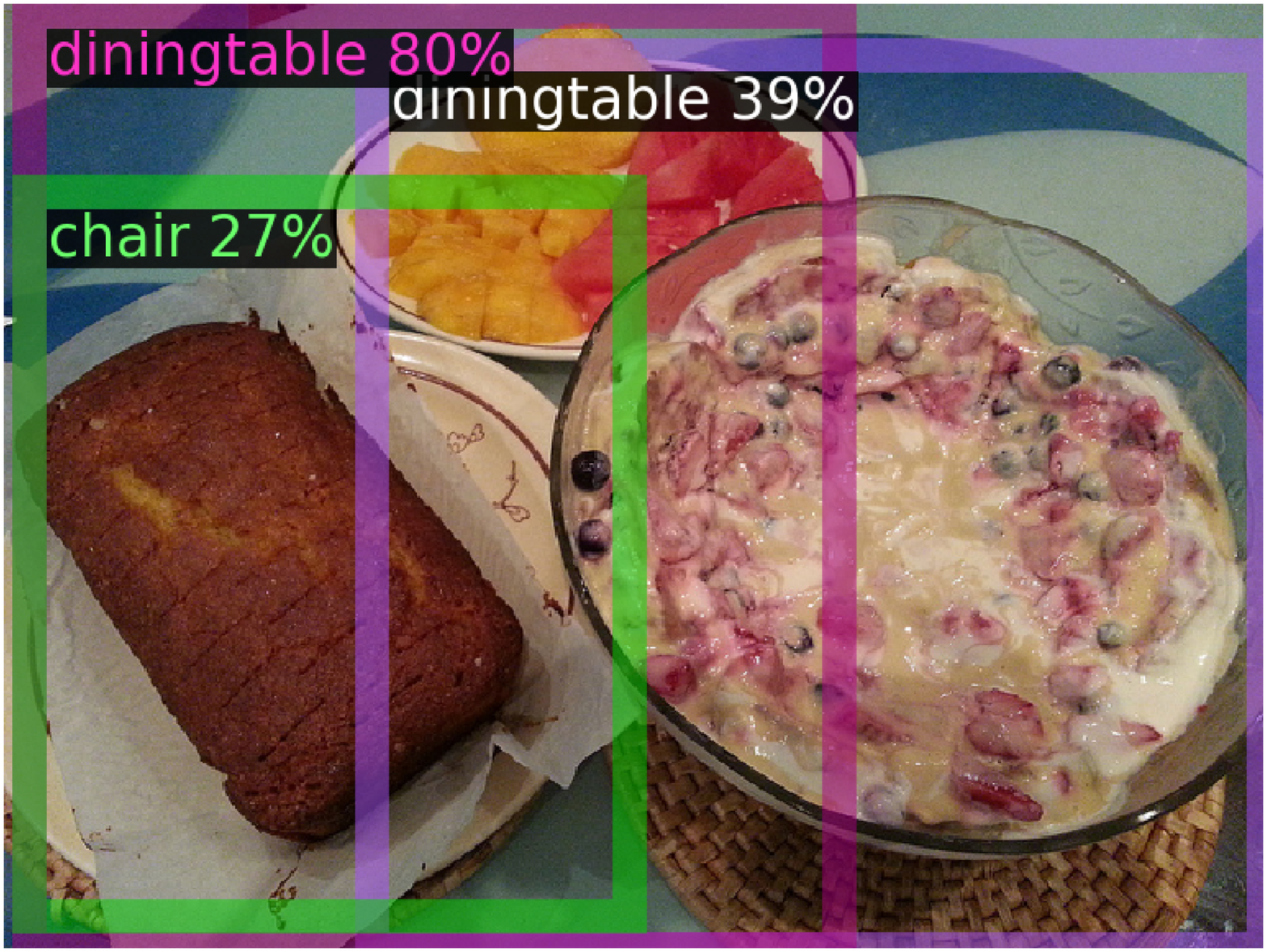}}
\end{minipage}
\hfill
\begin{minipage}[b]{0.49\linewidth}
  \centering
  \centerline{\includegraphics[width=\linewidth]{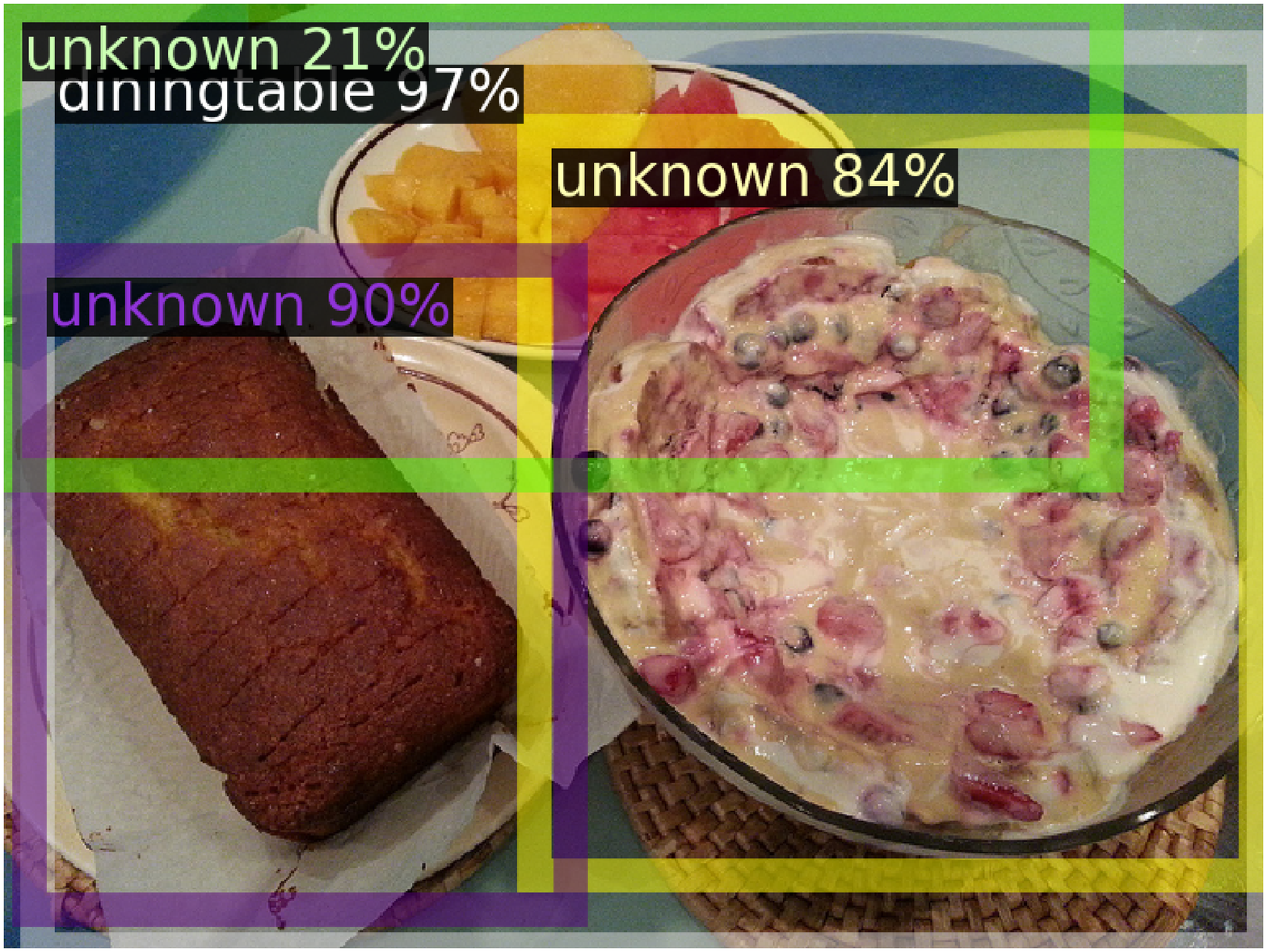}}
\end{minipage}
\begin{minipage}[b]{0.49\linewidth}
  \centering
  \centerline{\includegraphics[width=\linewidth]{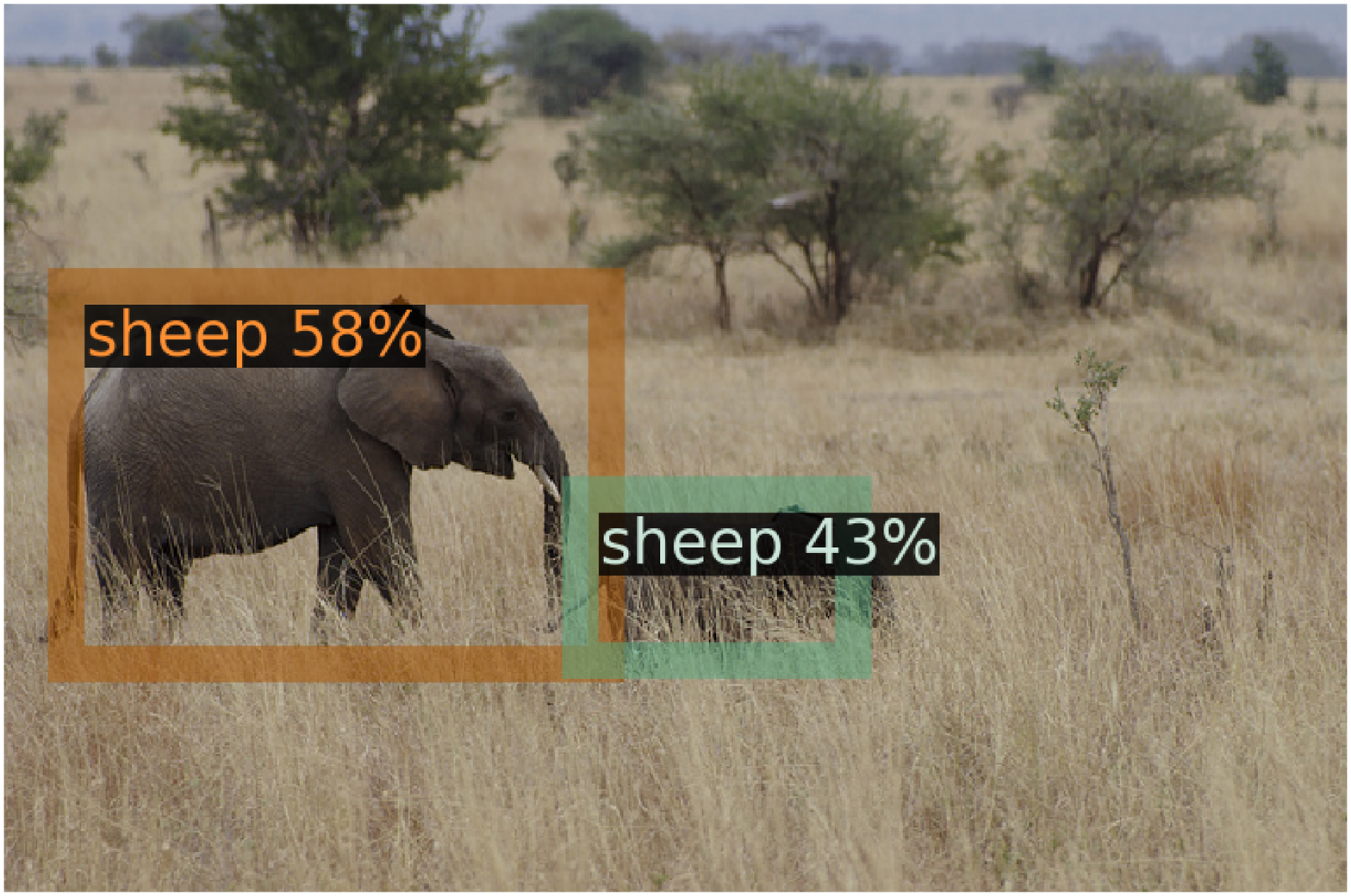}}
  \centerline{(a) ORE*}\medskip
\end{minipage}
\hfill
\begin{minipage}[b]{0.49\linewidth}
  \centering
  \centerline{\includegraphics[width=\linewidth]{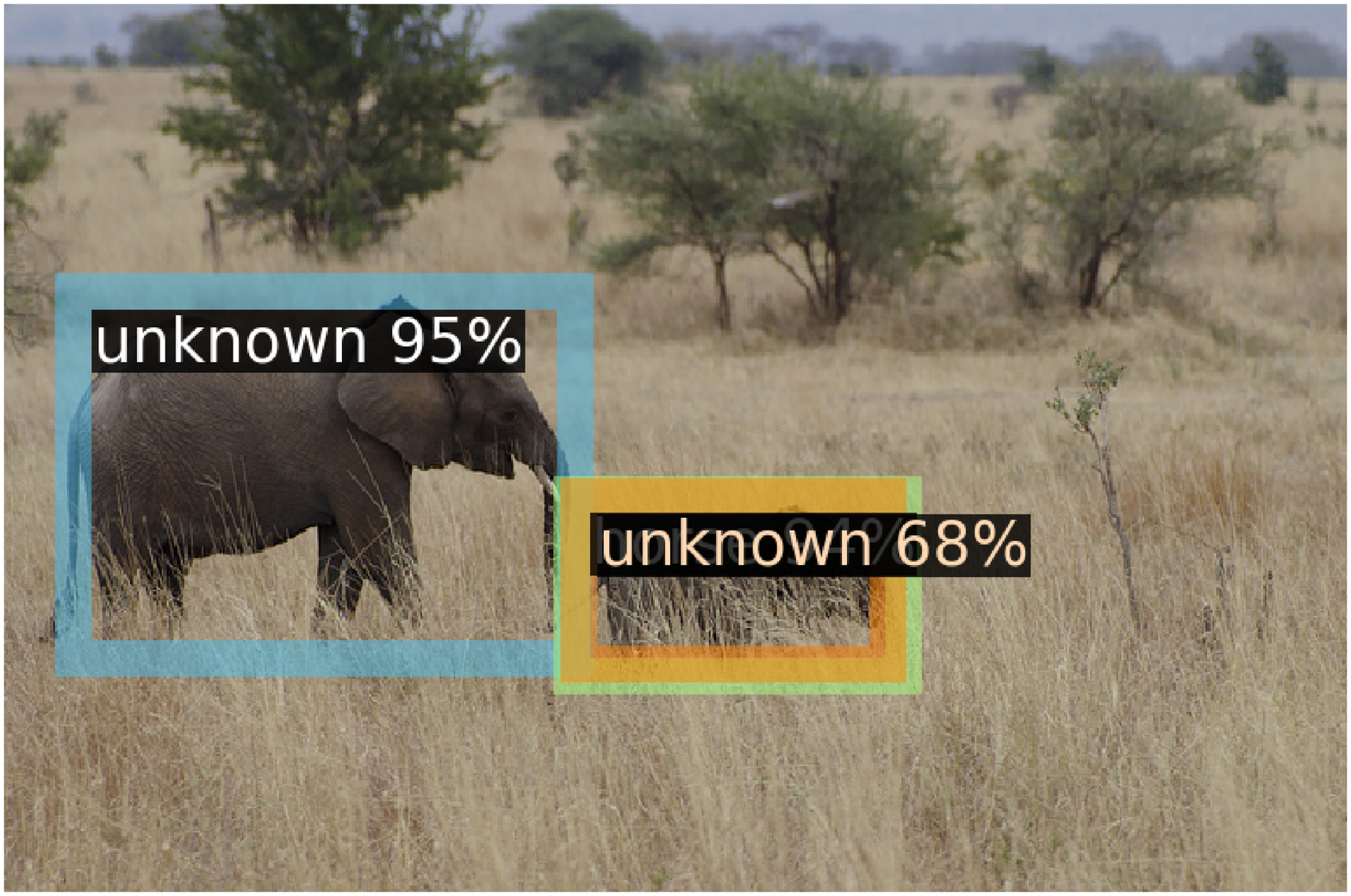}}
  \centerline{(b) Ours}\medskip
\end{minipage}
\caption{Qualitative results of different detectors}
\label{fig:visual}
\end{figure}

\section{CONCLUSION}
In this paper, we propose a novel and efficient solution for Open World Object Detection (OWOD). We are the first to try to incorporate a prototype ideology into the OWOD problem and achieve better performance than the recently proposed ORE on various metrics. Two constraints and a cosine classifier are introduced into our proposed method to avoid overlap between known and unknown class distributions as much as possible. Comprehensive experiments performed on Pascal VOC and MSCOCO demonstrate the effectiveness of our proposed method.

\vfill
\pagebreak

\bibliographystyle{IEEEbib}
\bibliography{strings,refs}

\end{document}